\PassOptionsToPackage{numbers}{natbib}
\documentclass{article}

\usepackage[preprint]{neurips_2024}
\usepackage[utf8]{inputenc} % allow utf-8 input
\usepackage[T1]{fontenc}    % use 8-bit T1 fonts
\usepackage{hyperref}       % hyperlinks
\usepackage{url}            % simple URL typesetting
\usepackage{booktabs}       % professional-quality tables
\usepackage{amsfonts}       % blackboard math symbols
\usepackage{nicefrac}       % compact symbols for 1/2, etc.
\usepackage{microtype}      % microtypography
\usepackage{xcolor}         % colors
\usepackage{enumitem}

\usepackage{natbib}

\usepackage{amsmath}
\usepackage{graphicx}
\usepackage{authblk}

\makeatletter
\renewcommand{\@noticestring}{}
\makeatother

\title{Not All Correct Answers Are Equal: Why Your Distillation Source Matters}

\begin{document}
\makeatletter
\renewcommand\@fnsymbol[1]{\ifcase#1\or 1\else\@arabic{#1}\fi}
\makeatother

\author{
  Xiaoyu Tian,\quad Yunjie Ji,\quad  Haotian Wang,\quad Shuaiting Chen,\\[0.3em]
  Sitong Zhao,\quad Yiping Peng,\quad Han Zhao,\quad Xiangang Li
}

\affil{
    \raisebox{-0.4em}{\includegraphics[height=1.5em]{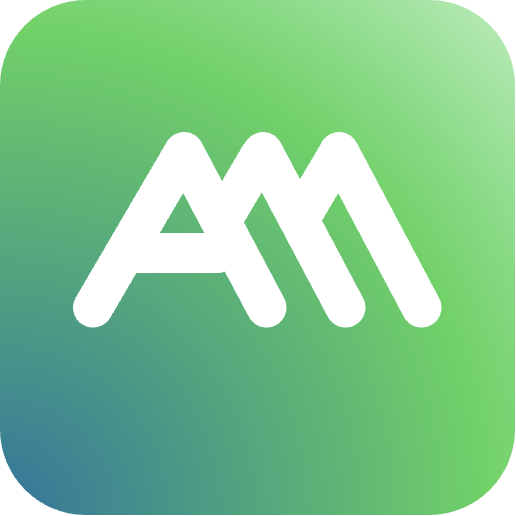}}
    \hspace{0.2em}a-m-team\thanks{The a-m-team is an internal team at Beike (Ke.com), dedicated to exploring AGI technology.}
}
\date{}

\maketitle

\vspace{-2.5em}
\begin{abstract}
\noindent Distillation has emerged as a practical and effective approach to enhance the reasoning capabilities of open-source language models. In this work, we conduct a large-scale empirical study on reasoning data distillation by collecting verified outputs from three state-of-the-art teacher models—AM-Thinking-v1, Qwen3-235B-A22B, and DeepSeek-R1—on a shared corpus of 1.89 million queries. We construct three parallel datasets and analyze their distributions, revealing that AM-Thinking-v1-distilled data exhibits greater token length diversity and lower perplexity. Student models trained on each dataset are evaluated on reasoning benchmarks including AIME2024, AIME2025, MATH500, and LiveCodeBench. The model distilled from AM-Thinking-v1 consistently achieves the best performance (e.g., 84.3 on AIME2024, 72.2 on AIME2025, 98.4 on MATH500, and 65.9 on LiveCodeBench) and demonstrates adaptive output behavior—producing longer responses for harder tasks and shorter ones for simpler tasks. These findings highlight the value of high-quality, verified reasoning traces. We release the AM-Thinking-v1 and Qwen3-235B-A22B distilled datasets to support future research on open and high-performing reasoning-oriented language models. The datasets are publicly available on Hugging Face\footnote{\href{https://huggingface.co/datasets/a-m-team/AM-Thinking-v1-Distilled}{https://huggingface.co/datasets/a-m-team/AM-Thinking-v1-Distilled}.}.
\end{abstract}

\vspace{-1.5em}
\begin{figure}[ht]
    \centering
    \includegraphics[width=1\linewidth]{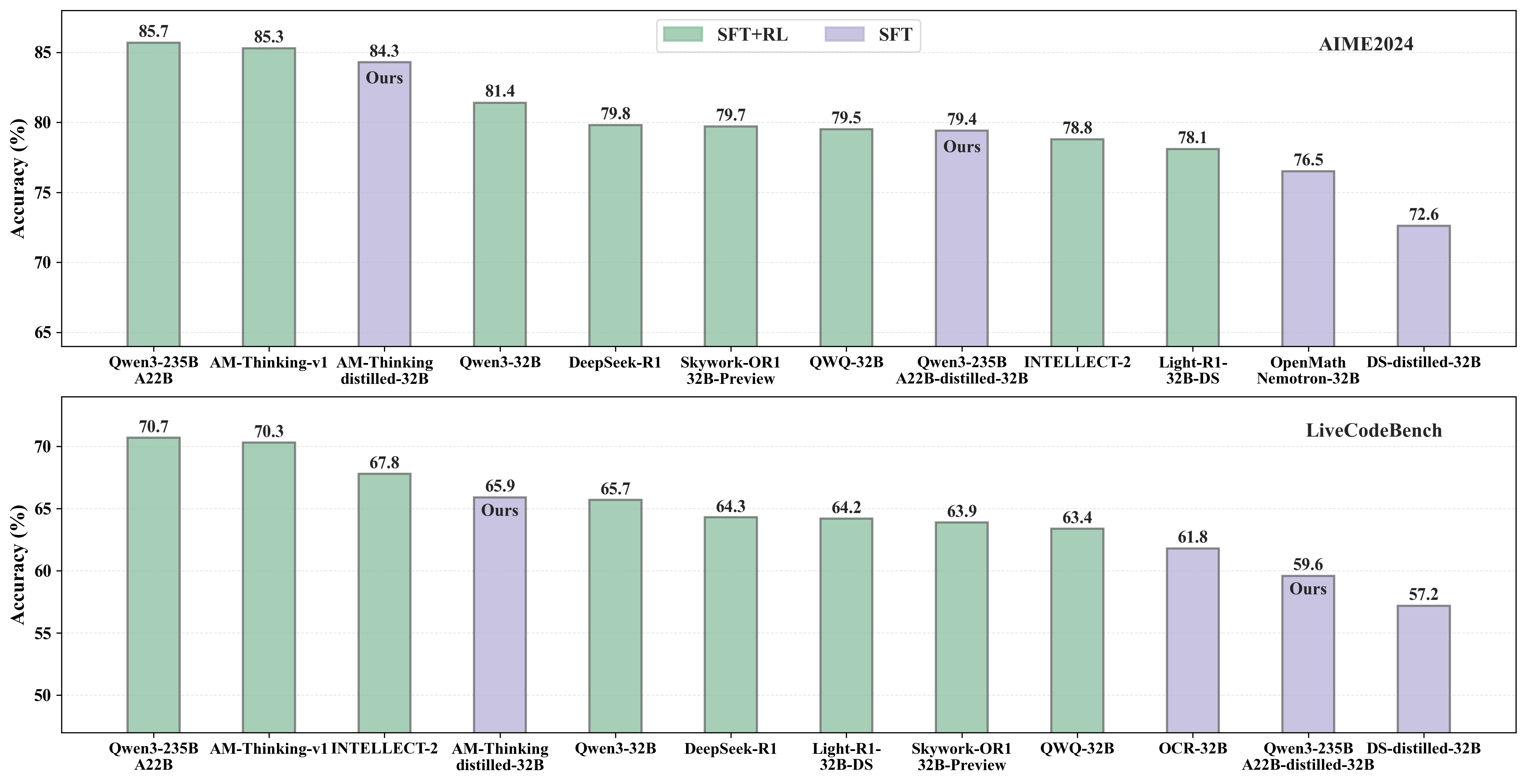}
    \vspace{-0.7em}
    \caption{Open-source model benchmarks on AIME2024/LiveCodeBench\protect\footnotemark.}
    \label{fig:hist_benchmark}
\end{figure}
\footnotetext{Due to the existence of multiple versions of LiveCodeBench, there may be small fluctuations in the scores.}

\clearpage
\section{Introduction}

Recent work has demonstrated the effectiveness and efficiency of distillation-based training for enhancing the reasoning ability of Large Language Models (LLMs)~\citep{deepseekai2025deepseekr1incentivizingreasoningcapability, tian2025deepdistillenhancingllmreasoning, qwen3}. By transferring reasoning traces from stronger teacher models, distilled data enables smaller or open-source models to achieve significant improvements on challenging tasks such as mathematics, coding, and scientific reasoning.

Building on this line of research, we systematically distilled reasoning data from three state-of-the-art models: DeepSeek-R1~\citep{deepseekai2025deepseekr1incentivizingreasoningcapability}, Qwen3-235B-A22B~\citep{qwen3}, and AM-Thinking-v1~\citep{ji2025amthinkingv1advancingfrontierreasoning}. For each of approximately 1.89 million identical queries, we collected full chain-of-thought responses from all three models, resulting in three parallel large-scale datasets. This unique setup allows for a direct comparison of reasoning styles and data distributions across leading models.

We carefully processed and cleaned all three datasets, including thorough deduplication, strict filtering, and contamination removal. We further analyzed the data distributions and content diversity to provide a comprehensive understanding of the strengths and characteristics of each distillation source.

Our experiments show that models trained with data distilled from AM-Thinking-v1 achieve particularly strong performance. On challenging reasoning benchmarks, such as AIME2024~\citep{maa_aime_2024} (84.3), AIME2025~\citep{ye2025aimepreview} (72.2), MATH500~\citep{lightman2023letsverifystepstep} (98.4), and LiveCodeBench~\citep{jain2024livecodebench} (65.9), the AM-Thinking-v1 distilled model consistently outperforms those trained on Qwen3-235B-A22B or DeepSeek-R1 data. Moreover, our analysis reveals that the AM-Thinking-v1 distilled model exhibits an adaptive generation length: producing longer responses on harder tasks (e.g., AIME, LiveCodeBench), and shorter ones on simpler datasets (e.g., MATH500). This behavior aligns with the token-level distribution of the AM-distilled dataset, which contains both short and long responses more frequently than the other two sources.

These results highlight the practical value of large-scale, high-quality reasoning data distillation, for improving open-source LLMs. To promote further progress in the field, we release both the AM-Thinking-v1 and Qwen3-235B-A22B distilled datasets\footnote{Note that since the distillation data of DeepSeek-R1 is easily accessible, we only release the distillation data of AM-Thinking-v1 and Qwen3-235B-A22B.}. We hope our work provides valuable resources and insights for the open-source community, enabling more effective reasoning-focused model development and contributing to the broader progress of reasoning research.

\section{Data}

This section first introduces the data preprocessing and distillation pipeline used to construct our training corpus, and then presents a detailed analysis of the resulting datasets in terms of distribution, length, and quality.

\subsection{Data Collection and Query Processing}

To support robust and comprehensive model training, we constructed a large-scale training corpus by aggregating data from a diverse set of publicly available open-source corpora. These corpora span a broad range of NLP tasks, including mathematical reasoning, code generation, scientific reasoning, instruction following, multi-turn dialogue, and general reasoning. For downstream analysis and targeted data processing, each data source was systematically assigned to a specific task category.

\paragraph{Training Data Categories} The aggregated training data were classified as follows:

\begin{itemize}
    \item \textbf{Mathematical Reasoning}: Datasets requiring advanced numerical reasoning and multi-step logic, such as OpenR1-Math-220k~\citep{openr1}, Big-Math-RL-Verified~\citep{albalak2025bigmathlargescalehighqualitymath}, NuminaMath~\citep{numina_math_datasets}, among others.
    \item \textbf{Code Generation}: Datasets aimed at enhancing code synthesis and programmatic problem-solving abilities, including PRIME~\citep{yuan2024implicitprm}, DeepCoder~\citep{deepcoder2025}, KodCode~\citep{xu2025kodcode}.
    \item \textbf{Scientific Reasoning}: Datasets emphasizing reasoning within the natural sciences, such as task\_mmmlu~\citep{wang2022supernaturalinstructionsgeneralizationdeclarativeinstructions}, chemistryQA~\citep{microsoft2021chemistryqa}, and LOGIC-701~\citep{hivaze_logic701_2023}.
    \item \textbf{Instruction Following (IF)}: Data focused on instruction comprehension and faithful execution, including Llama-Nemotron-Post-Training-Dataset~\citep{nvidia2025llama}, tulu-3-sft-mixture~\citep{lambert2024tulu3}, if-eval-like, and AutoIF.
    \item \textbf{Multi-turn Conversation}: Corpora curated to train dialogue agents on contextually coherent and consistent multi-turn interactions, such as InfinityInstruct~\citep{InfinityInstruct2024}, OpenHermes-2.5~\citep{OpenHermes_2_5}, and ultra\_chat~\citep{ding2023enhancing}.
    \item \textbf{General Reasoning}: Datasets covering diverse open-ended reasoning and general knowledge tasks, including evol~\citep{wizardlm_evol_instruct_70k}, open\_orca~\citep{OpenOrca}, flan~\citep{goodson2023huggyflan}.
\end{itemize}

\paragraph{Query Preprocessing}  
To guarantee the reliability of subsequent model training, we applied rigorous multi-stage preprocessing to the raw queries:

\begin{enumerate}
    \item \textbf{Deduplication}: Exact duplicate queries (identical text) were removed.
    \item \textbf{Filtering}: 
        \begin{itemize}
            \item Queries with a high Unicode character ratio were discarded to eliminate corrupted or meaningless samples.
            \item Incomplete or empty queries were excluded.
            \item Instances containing URLs or tabular structures were filtered out to reduce noise and hallucination risk.
        \end{itemize}
    \item \textbf{Decontamination}: To mitigate data contamination, especially regarding the core evaluation set (e.g., AIME2024~\citep{maa_aime_2024}), we conducted both exact match filtering and semantic deduplication. The latter leveraged the bge-m3 embedding model~\citep{bge-m3} to compute semantic similarity, removing queries exceeding a threshold of 0.9 with respect to the evaluation set.
\end{enumerate}

\subsection{Data Distilling}

After preprocessing, we performed large-scale data distillation to further enhance the quality of our training corpus.

\paragraph{Distillation Framework}
For each preprocessed query, we adopted an incremental distillation strategy using three state-of-the-art models: AM-Thinking-v1~\citep{ji2025amthinkingv1advancingfrontierreasoning}, Qwen3-235B-A22B~\citep{qwen3}, and DeepSeek-R1~\citep{deepseekai2025deepseekr1incentivizingreasoningcapability}. Each query was independently distilled by these three models. For every model, the distillation process was repeated on the same query until the generated response satisfied the verification criterion (i.e., the verification score $\geq 0.9$). Consequently, each query yielded up to three high-quality distilled outputs, corresponding to the three models, with each output refined iteratively until it passed the automatic verification.

\paragraph{Automatic Verification and Scoring}
To ensure the reliability and correctness of the distilled data, we employed automatic verification procedures tailored for each data category, assigning a verification score (\textit{verify\_score}) to every model-generated response:

\begin{itemize}
    \item \textbf{Mathematical Reasoning}: Responses were verified using a two-stage process—first with Math-Verify\footnote{\url{https://github.com/huggingface/Math-Verify}}, and, if necessary, subsequently with Qwen2.5-7B-Instruct~\citep{qwen2, qwen2.5}. Each result was assigned a binary verification score.
    \item \textbf{Code Generation}: Each code response was validated in a sandbox environment using up to 10 test cases (assert and input-output for Python; input-output for C++), with the verification score reflecting the pass rate.
    \item \textbf{Scientific Reasoning}: The similarity between predicted and reference answers was assessed using Qwen2.5-7B-Instruct~\citep{qwen2, qwen2.5}, yielding a normalized score.
    \item \textbf{Instruction Following}: Responses were verified with the ifeval validator, supplementing missing constraints with Qwen2.5-72B-Instruct~\citep{qwen2, qwen2.5}. The mean pass rate over all constraints was taken as the verification score.
    \item \textbf{Multi-turn Conversations and General Reasoning}: Decision-Tree-Reward-Llama-3.1-8B~\citep{rlhflow2025decisiontree} was used to evaluate coherence, correctness, and helpfulness, which were aggregated into a normalized composite score.
\end{itemize}

A unified verification score threshold of 0.9 was used across all data categories.

\paragraph{Quality Assurance Measures}
To further enhance data quality, we introduced several additional validation and filtering strategies:
\begin{itemize}
    \item \textbf{Perplexity-based Filtering}: We computed perplexity scores using a strong 32B language model~\citep{zhao20251}, with each model employing a different threshold. Notably, responses distilled from AM-Thinking-v1 demonstrated the lowest perplexity among the three models.
    \item \textbf{High-Frequency Ngram Filtering}: 20-token ngrams occurring more than 20 times were identified and removed to reduce template-like redundancy.
    \item \textbf{Logical and Structural Validation}: Checks included ensuring an even number of dialogue turns for conversation data, explicit presence of both reasoning (“think”) and answer segments in each sample.
\end{itemize}

Ultimately, this process yielded a comprehensive dataset of 1.89 million queries, each paired with high-quality, verified responses distilled from all three models.

\subsection{Data Analysis}
\label{sec_data_analysis}

\begin{figure}[ht]
    \centering
    \includegraphics[width=1\linewidth]{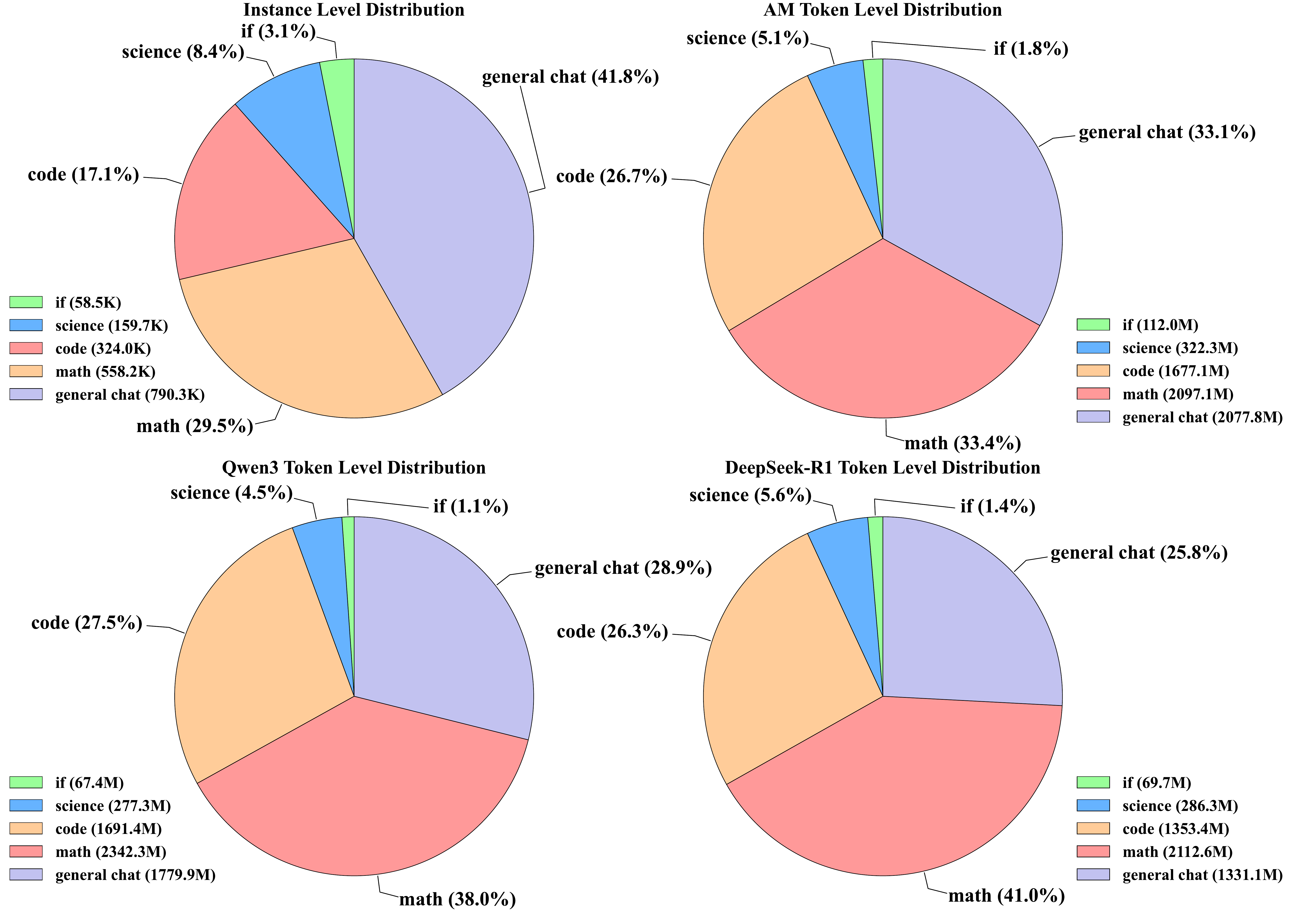}
    \caption{Instance-level and token-level output distributions are analyzed for AM-Thinkin-v1, Qwen3-235B-A22B, and DeepSeek-R1. The general chat includes both multi-turn conversations and other types of data.}
    \label{fig:4_combine_distribution}
\end{figure}

\begin{figure}[ht]
    \centering
    \includegraphics[width=1\linewidth]{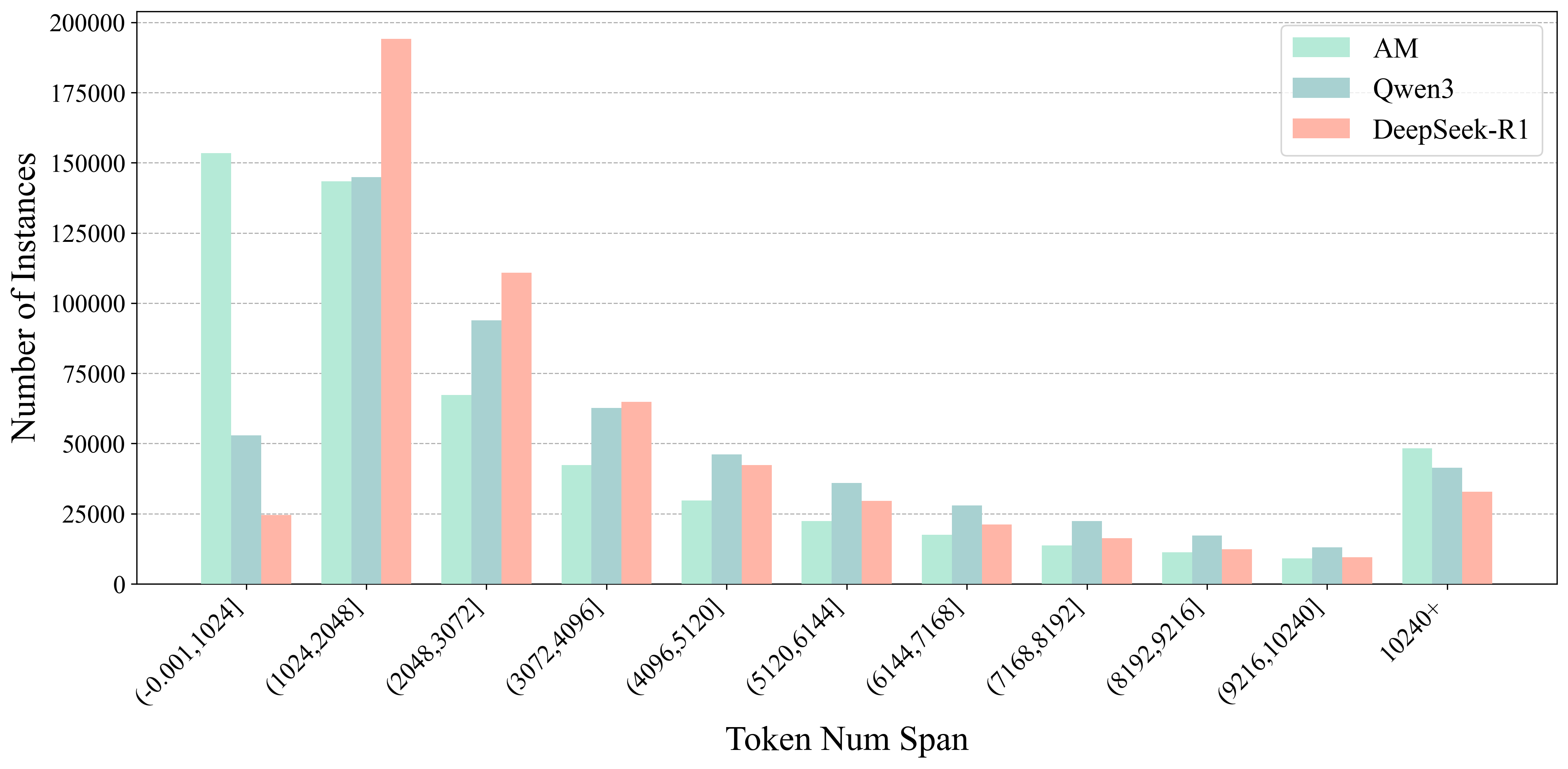}
    \caption{Token span distribution of instances for AM-Thinking-v1, Qwen3-235B-A22B, and DeepSeek-R1 on math.}
    \label{fig:math_length_dist_plot}
\end{figure}

\begin{figure}[ht]
    \centering
    \includegraphics[width=0.65\linewidth]{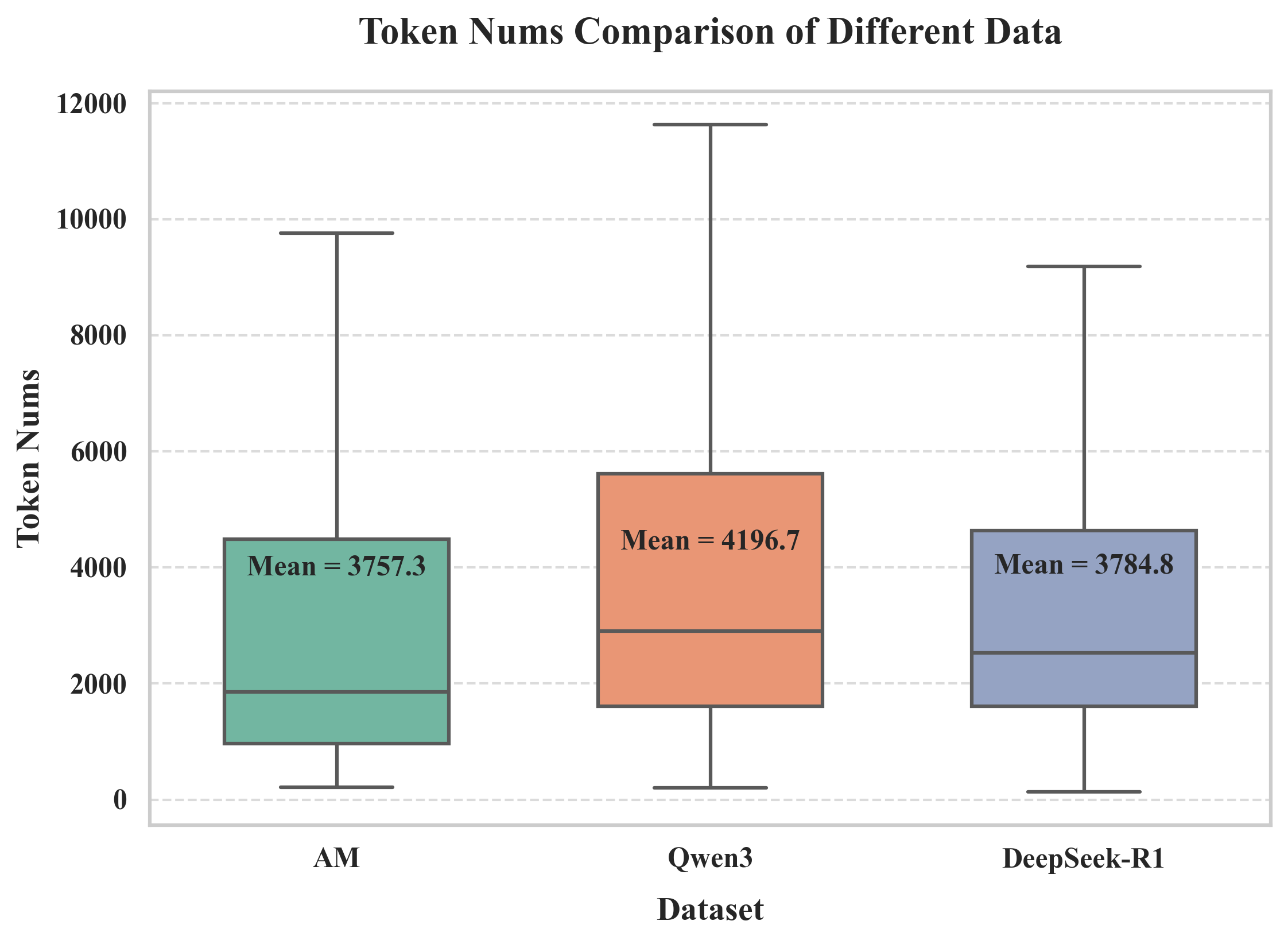}
    \caption{Token count distributions for AM-Thinking-v1, Qwen3-235B-A22B, and DeepSeek-R1 datasets. Box plots show the distribution of token numbers, with means labeled. Qwen3-235B-A22B has the highest average token count, followed by DeepSeek-R1 and AM-Thinking-v1.}
    \label{fig:boxplot_for_length_math}
\end{figure}

\begin{figure}[ht]
    \centering
    \includegraphics[width=0.65\linewidth]{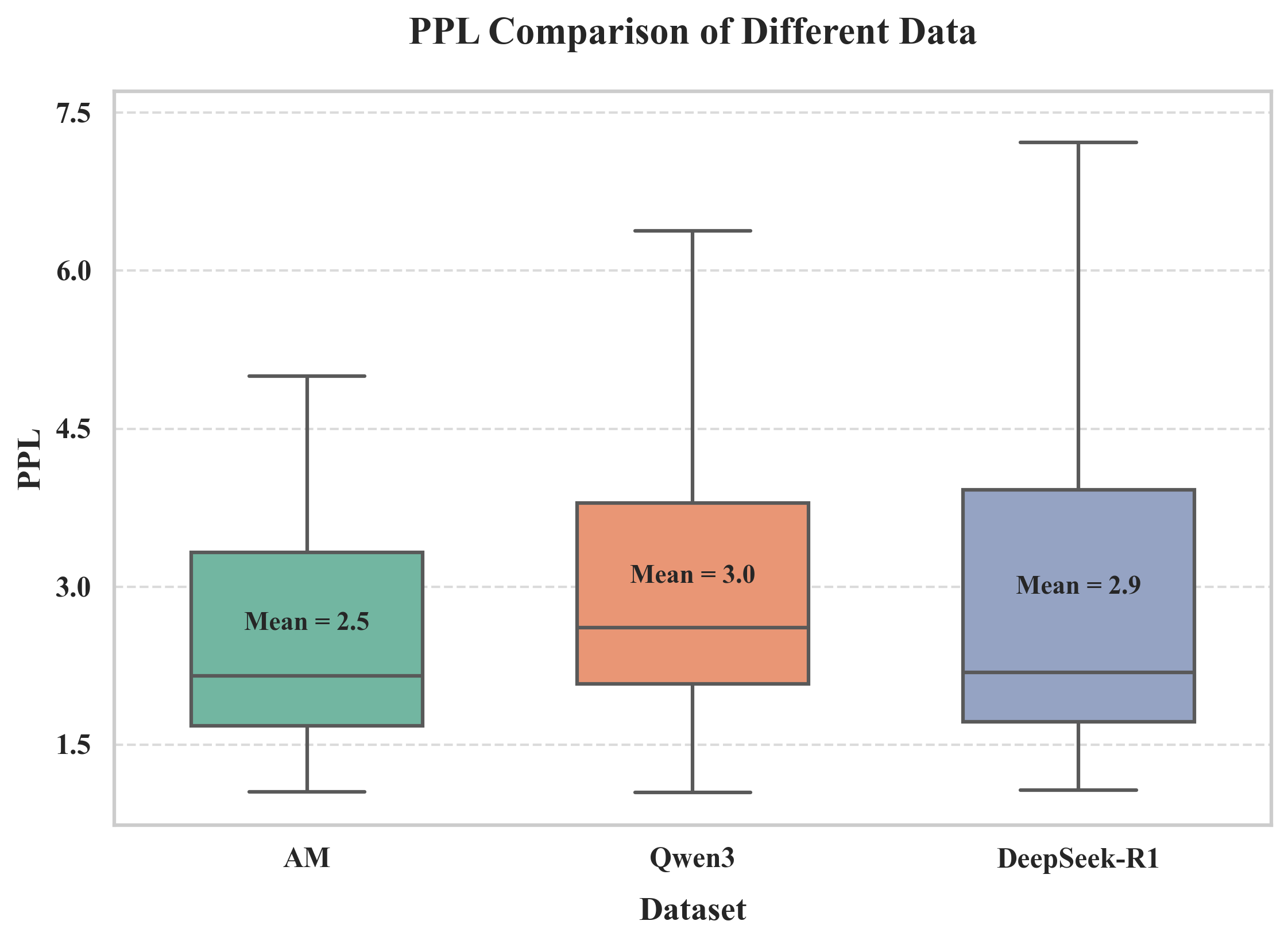}
    \caption{Perplexity (PPL) distributions for AM-Thinking-v1, Qwen3-235B-A22B, and DeepSeek-R1 datasets. Box plots show PPL distributions, with means labeled. AM-Thinking-v1 achieves the lowest mean PPL, indicating better overall quality.}
    \label{fig:boxplot_for_ppl}
\end{figure}

We conduct a detailed analysis of the training data distilled from three different large-scale models: AM-Thinking-v1~\citep{ji2025amthinkingv1advancingfrontierreasoning}, Qwen3-235B-A22B~\citep{qwen3}, and DeepSeek-R1~\citep{deepseekai2025deepseekr1incentivizingreasoningcapability}. This comparative analysis covers the instance-level and token-level output distributions, token length characteristics, and perplexity (PPL) distributions, providing insight into the data quality and structural tendencies of each dataset.

Figure~\ref{fig:4_combine_distribution} shows the output distribution of these datasets at both the instance and token levels. The instance-level distribution (top-left) reveals that this dataset includes a high proportion of general chat (41.8\%), followed by math (29.5\%) and code (17.1\%). In contrast, Qwen3-235B-A22B and DeepSeek-R1 token-level distributions (bottom charts) show a more pronounced focus on math (38.0\% and 41.0\% respectively), with general chat and code sharing similar proportions. Notably, AM-Thinking-v1's token-level distribution (top-right) also emphasizes math (33.4\%), albeit to a lesser extent than DeepSeek-R1. Science and instruction-following (IF) data make up a minor share across all datasets.

Further, Figure~\ref{fig:math_length_dist_plot} presents the token span distribution specifically for math instances. It demonstrates that AM-Thinking-v1's math data exhibits a highly dispersed distribution—many short sequences (under 1024 tokens) and a substantial portion of very long sequences (10240+ tokens). This reflects the token distribution characteristics under a fixed query set, where AM-Thinking-v1 tends to produce both short and very long responses more frequently. In contrast, Qwen3-235B-A22B's data generally exhibits longer token spans, indicating a tendency toward producing longer responses. DeepSeek-R1's token spans are mostly concentrated between 1k and 8k tokens, showing a moderate range of response lengths.

Box plots in Figure~\ref{fig:boxplot_for_length_math} provide further insight into the average token length per instance. The Qwen3-235B-A22B dataset has the highest mean token count (4196.7), followed by DeepSeek-R1 (3784.8) and AM-Thinking-v1 (3757.3). This aligns with the histogram observation that Qwen3-235B-A22B emphasizes longer instances, whereas AM-Thinking-v1 covers a broader range of lengths, including extremely short and long sequences.

Finally, Figure~\ref{fig:boxplot_for_ppl} compares the perplexity (PPL) across the datasets. Perplexity is a key measure of language model performance, with lower values indicating better quality. Among the three datasets, AM-Thinking-v1 achieves the lowest mean PPL (2.5), suggesting that its distilled outputs are generally of higher quality. DeepSeek-R1 (mean PPL = 2.9) performs slightly better than Qwen3 (mean PPL = 3.0), highlighting the relatively strong performance of AM-distilled data in terms of perplexity.

\section{Experiments}

\subsection{Training Configuration}

Training is conducted based on the Qwen2.5-32B~\citep{qwen2, qwen2.5} base model. As suggested by~\citep{tian2025deepdistillenhancingllmreasoning, ji2025amthinkingv1advancingfrontierreasoning}, all three models are trained with a learning rate of 8e-5, a maximum sequence length of 32k (using sequence packing), and a global batch size of 64 for 2 epochs. Samples longer than 32k tokens are excluded. We apply cosine warmup with 5\% of total steps, and the learning rate decays to zero thereafter. For multi-turn dialogues, only the final response containing the reasoning process is used as the training target to focus learning on reasoning.

\subsection{Benchmarks and Evaluation Setup}
\label{sec:eval_setup}
To rigorously assess our models' capabilities, we select a diverse suite of challenging benchmarks across mathematical reasoning, programming, and general chatbot performance:

\begin{itemize}
\item \textbf{AIME2024}~\citep{maa_aime_2024}: A high-difficulty dataset featuring 30 integer-answer questions from the 2024 American Invitational Mathematics Examination, designed to evaluate accurate mathematical problem-solving skills.
\item \textbf{AIME2025}~\citep{ye2025aimepreview}: Contains 30 new problems curated from the 2025 AIME Part I and Part II exams, offering a forward-looking benchmark for advanced mathematical reasoning.
\item \textbf{LiveCodeBench (LCB)}~\citep{jain2024livecodebench}: A dynamically evolving, contamination-free code generation benchmark. Tasks are aggregated from platforms like LeetCode, Codeforces, and AtCoder. In line with prior works such as Qwen3~\cite{qwen3}, we use a snapshot of queries submitted between October 2024 and February 2025.
%\item \textbf{Arena-Hard}~\citep{arenahard2024}: A curated benchmark built from real-world chatbot interactions collected via Chatbot Arena. Responses are evaluated using GPT-4-Turbo-1106~\citep{openai2024reasoning} in a pairwise setting to determine comparative model performance.

\item \textbf{MATH500}~\citep{lightman2023letsverifystepstep}: A benchmark consisting of 500 challenging math word problems designed to evaluate the problem-solving abilities of large language models. It covers diverse mathematical domains such as algebra, geometry, calculus, and number theory, requiring models to perform multi-step reasoning and symbolic manipulation.
\end{itemize}

All benchmarks were evaluated under uniform conditions. The generation length was capped at 49,152 tokens. For stochastic decoding, we consistently adopted a temperature of 0.6 and top-p of 0.95 across applicable tasks.

Response sampling was tailored to the nature of each benchmark:
\begin{itemize}
\item \textbf{AIME2024 and AIME2025}: For each question, we generated 64 outputs to estimate pass@1 accuracy.
\item \textbf{LiveCodeBench}: We sampled 16 completions per prompt to compute pass@1.
%\item \textbf{Arena-Hard}: Each prompt was answered once, and model outputs were judged via pairwise comparison using GPT-4-Turbo.

\item \textbf{MATH500}: Each prompt was answered once, and we sampled 4 times to compute pass@1.
\end{itemize}

Followed by AM-Thinking-v1\citep{ji2025amthinkingv1advancingfrontierreasoning}, a unified system prompt was employed across all tasks to standardize output format and encourage reasoning:

\begin{quote}
\texttt{You are a helpful assistant. To answer the user's question, you first think about the reasoning process and then provide the user with the answer. The reasoning process and answer are enclosed within <think> </think> and <answer> </answer> tags, respectively, i.e., <think> reasoning process here </think> <answer> answer here </answer>.}
\end{quote}

User prompts were benchmark-specific:
\begin{itemize}
\item \textbf{AIME tasks  and MATH500}: A supplementary instruction was added: \texttt{Let's think step by step and output the final answer within \textbackslash box{}.}
\item \textbf{LiveCodeBench}: Original task prompts were used without any alterations.
\end{itemize}

\section{Results and Analysis}
% v5, 2024.10–2025.02
% \begin{table}[htbp]
%     \caption{Comparison across reasoning benchmarks using distilled data from different teacher models.}
%     \label{tab:final_results_top3}
%     \centering
%     \renewcommand{\arraystretch}{1.3}
%     \begin{tabular}{l c c c}
%         \hline
%         & $\text{AM-Thinking-v1}_{\text{Distilled}}$ 
%         & $\text{Qwen3-235B-A22B}_{\text{Distilled}}$
%         & $\text{DeepSeek-R1}_{\text{Distilled}}$ \\
%         \hline
%         AIME 2024 & 84.3 & 79.4 & 70.9 \\
%         AIME 2025 & 72.2 & 62.2 & 52.8 \\
%         MATH500 & 98.4 & 93.9 &  \\
%         LiveCodeBench\vspace{-1ex} & 65.9 & 59.6 &  \\
%         \multicolumn{4}{l}{\scriptsize (v5, 2024.10--2025.02)} \\
%         Arena‑Hard & 89.4 & 93.6 &  \\
%         \hline
%     \end{tabular}
% \end{table}
\begin{table}[htbp]
    \caption{Comparison across reasoning benchmarks using distilled data from different teacher models.}
    \label{tab:final_results}
    \centering
    \renewcommand{\arraystretch}{1.3}
    \begin{tabular}{l c c c}
        \hline
        & $\text{AM-Thinking-v1}_{\text{Distilled}}$ 
        & $\text{Qwen3-235B-A22B}_{\text{Distilled}}$
        & $\text{DeepSeek-R1}_{\text{Distilled}}$ \\
        \hline
        AIME2024 & \textbf{84.3} & 79.4 & 70.9 \\
        AIME2025 & \textbf{72.2} & 62.2 & 52.8 \\
        MATH500 & \textbf{98.4} & 93.9 & 95.8 \\
        LiveCodeBench\vspace{-1ex} & \textbf{65.9} & 59.6 & 57.0 \\
        \multicolumn{4}{l}{\scriptsize (v5, 2024.10--2025.02)} \\
        \hline
    \end{tabular}
\end{table}

% \begin{table}[htbp]
%     \caption{Average generation length (tokens per sample) across reasoning benchmarks.}
%     \label{tab:avg_gen_length_top3}
%     \centering
%     \renewcommand{\arraystretch}{1.3}
%     \begin{tabular}{l c c c}
%         \hline
%         & $\text{AM-Thinking-v1}_{\text{Distilled}}$ 
%         & $\text{Qwen3-235B-A22B}_{\text{Distilled}}$
%         & $\text{DeepSeek-R1}_{\text{Distilled}}$ \\
%         \hline
%         AIME 2024 & 15273.8 & 13516.4 & 11853.5 \\
%         AIME 2025 & 18199.2 & 16975.7 & 13495.9 \\
%         MATH500 & 4023.8 & 6429.4 &  \\
%         LiveCodeBench\vspace{-1ex} & 23426.9 & 13576.7 &  \\
%         \multicolumn{4}{l}{\scriptsize (v5, 2024.10--2025.02)} \\
%         Arena‑Hard & 5697.6 & 4023.8 &  \\
%         \hline
%     \end{tabular}
% \end{table}
\begin{table}[htbp]
    \caption{Average generation length (tokens per sample) across reasoning benchmarks.}
    \label{tab:avg_gen_length}
    \centering
    \renewcommand{\arraystretch}{1.3}
    \begin{tabular}{l c c c}
        \hline
        & $\text{AM-Thinking-v1}_{\text{Distilled}}$ 
        & $\text{Qwen3-235B-A22B}_{\text{Distilled}}$
        & $\text{DeepSeek-R1}_{\text{Distilled}}$ \\
        \hline
        AIME2024 & 15273.8 & 13516.4 & 11853.5 \\
        AIME2025 & 18199.2 & 16975.7 & 13495.9 \\
        MATH500 & 3495.7 & 6429.4 & 3613.0 \\
        LiveCodeBench\vspace{-1ex} & 23426.9 & 13576.7 & 30731.0 \\
        \multicolumn{4}{l}{\scriptsize (v5, 2024.10--2025.02)} \\
        \hline
    \end{tabular}
\end{table}

We evaluate the models on the reasoning benchmarks described in Section~
\ref{sec:eval_setup}, using models trained with data distilled from AM-Thinking-v1, Qwen3-235B-A22B, and DeepSeek-R1. The evaluation results are presented in Table~\ref{tab:final_results}.

As shown in Table~\ref{tab:final_results}, the model distilled from AM-Thinking-v1 consistently achieves the highest accuracy across all benchmarks. On the more challenging math tasks, AIME2024 and AIME2025, it attains scores of 84.3 and 72.2, respectively, outperforming the Qwen3- and DeepSeek-distilled models by a considerable margin. It also leads on MATH500 (98.4) and LiveCodeBench (65.9), indicating broad generalization across both mathematical and code-based reasoning tasks.

To better understand model behavior, we analyze the average generation length per sample across benchmarks (Table~\ref{tab:avg_gen_length}). Interestingly, the $\text{AM-Thinking-v1}_{\text{Distilled}}$ model produces notably longer outputs on more complex tasks: 15273.8 and 18199.2 tokens for AIME2024 and AIME2025, respectively, and 23426.9 for LiveCodeBench. In contrast, on the simpler MATH500 benchmark, its average generation length (3495.7) is shorter than that of the $\text{Qwen3-235B-A22B}_{\text{Distilled}}$ model. This adaptive generation pattern suggests that the AM-distilled model can better modulate its output length based on task complexity---generating more detailed solutions when needed while remaining concise on simpler problems. This aligns with our earlier analysis in Section~\ref{sec_data_analysis}, where the AM-Thinking-v1 distilled dataset exhibited a higher proportion of both short and long token sequences. The broader distribution of token lengths in the training data likely contributes to the model's improved ability to adjust its response length dynamically. Such length modulation is a desirable property in reasoning tasks, where overgeneration or undergeneration can negatively impact performance.

\begin{figure}[ht]
    \centering
    \includegraphics[width=1\linewidth]{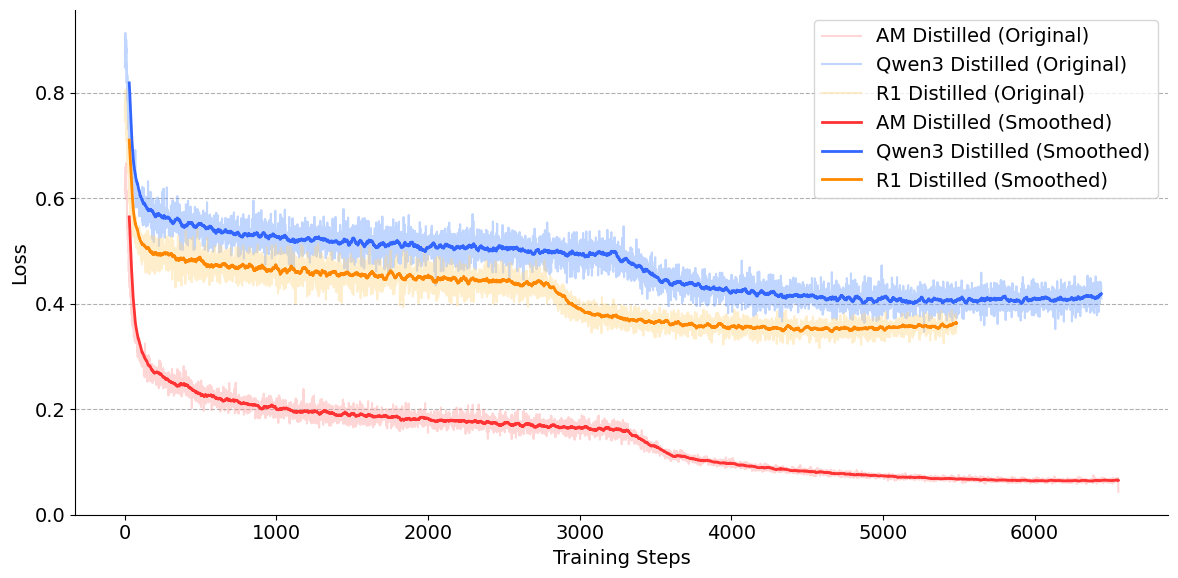}
    \caption{Loss curves of AM-Thinking-v1-Distilled, DeepSeek-R1-Distilled and Qwen3-235B-A22B-Distilled.}
    \label{fig:loss_curve}
\end{figure}
We further compare training dynamics by examining the loss curves shown in Figure~\ref{fig:loss_curve}. The $\text{AM-Thinking-v1}_{\text{Distilled}}$ maintains a consistently lower training loss than its $\text{Qwen3-235B-A22B}_{\text{Distilled}}$ and $\text{DeepSeek-R1}_{\text{Distilled}}$ counterparts throughout the optimization process. This observation supports the notion that the AM-Thinking-v1 dataset provides more learnable, coherent, and high-quality supervision signals for the base model.

\section{Conclusion and Future Work}

In this work, we present a comprehensive empirical study on reasoning data distillation for open-source language models. Using three state-of-the-art teacher models—AM-Thinking-v1, Qwen3-235B-A22B, and DeepSeek-R1—we constructed a large-scale parallel corpus comprising 1.89 million verified reasoning samples. Through rigorous data preprocessing, verification scoring, and quality assurance, we ensured the construction of high-quality training data suitable for robust student model learning.

Empirical results across a diverse set of benchmarks, including AIME2024 (84.3), AIME2025 (72.2), MATH500 (98.4), and LiveCodeBench (65.9), demonstrate that models trained on AM-Thinking-v1-distilled data consistently achieve strong performance. 

To gain deeper insights into model behavior, we conducted detailed analysis of generation behavior and training dynamics. We observed that the AM-distilled model exhibits adaptive generation length—producing longer responses on harder tasks and shorter ones on easier benchmarks—indicating its capacity to adjust to task difficulty. This aligns with our earlier data analysis showing that AM-Thinking-v1-distilled data features a wide range of token lengths, providing stronger support for adaptive reasoning.

Looking ahead, a promising direction for future work is to further enhance these models using reinforcement learning techniques, such as Proximal Policy Optimization (PPO) or Generalized Group Relative Policy Optimization (GRPO), to enhance reasoning ability and alignment. We release the distilled datasets based on AM-Thinking-v1 and Qwen3-235B-A22B to support ongoing research in open and high-performing reasoning-oriented language models.

%%%%%%%%%%%%%%%%%%%%%%%%%%%%%%%%%%%%%%%%%%%%%%%%%%%%%%%%%%%%
\newpage
\bibliographystyle{unsrt}
\bibliography{reference}

%%% %%% %%% %%% %%% %%% %%% %%% %%% %%% %%% %%% %%% %%% %%%
%                         Appendix                        %
%%% %%% %%% %%% %%% %%% %%% %%% %%% %%% %%% %%% %%% %%% %%%
% \clearpage
% \appendix

% \section{Data Analysis}
% \label{data_analysis}

% \begin{figure}[ht]
%     \centering
%     \includegraphics[width=1\linewidth]{length_dist_plot.png}
%     \caption{Token span distribution of instances for AM-Thinking-v1, Qwen3-235B-A22B, and DeepSeek-R1 over the full dataset.}
%     \label{fig:token_distribution}
% \end{figure}

% \section{Appendix / supplemental material}

% Optionally include supplemental material (complete proofs, additional experiments and plots) in appendix.
% All such materials \textbf{SHOULD be included in the main submission.}

%%%%%%%%%%%%%%%%%%%%%%%%%%%%%%%%%%%%%%%%%%%%%%%%%%%%%%%%%%%%

\end{document}